\documentclass[letterpaper]{article} 
\usepackage{subcaption}
\usepackage{amsmath}
\usepackage{booktabs}
\usepackage{tabularx}
\usepackage{multirow}
\usepackage{aaai25}  
\usepackage{times}  
\usepackage{helvet}  
\usepackage{courier}  
\usepackage[hyphens]{url}  
\usepackage{graphicx} 
\usepackage{natbib}  
\usepackage{caption} 
\usepackage{algorithm}
\usepackage{algorithmic} 
\usepackage{graphicx}
\usepackage{listings}
\DeclareCaptionStyle{ruled}{labelfont=normalfont,labelsep=colon,strut=off} 
\floatstyle{ruled}
\newfloat{listing}{tb}{lst}{}
\floatname{listing}{Listing}
\title{Prompt-SID: Learning Structural Representation Prompt via Latent Diffusion \\for Single-Image Denoising}
\author{
    Huaqiu Li\thanks{These authors contributed equally.}, Wang Zhang$^{*}$, Xiaowan Hu, Tao Jiang, Zikang Chen, Haoqian Wang\thanks{Haoqian Wang is corresponding author.}
}
\affiliations{
    Tsinghua University\\

    lihq23@mails.tsinghua.edu.cn
}

\usepackage{bibentry}

\begin{document}

\maketitle

\begin{abstract}
Many studies have concentrated on constructing supervised models utilizing paired datasets for image denoising, which proves to be expensive and time-consuming. Current self-supervised and unsupervised approaches typically rely on blind-spot networks or sub-image pairs sampling, resulting in pixel information loss and destruction of detailed structural information, thereby significantly constraining the efficacy of such methods. In this paper, we introduce Prompt-SID, a \textbf{prompt}-learning-based \textbf{s}ingle \textbf{i}mage \textbf{d}enoising framework that emphasizes preserving of structural details. This approach is trained in a self-supervised manner using downsampled image pairs. It captures original-scale image information through structural encoding and integrates this prompt into the denoiser. To achieve this, we propose a structural representation generation model based on the latent diffusion process and design a structural attention module within the transformer-based denoiser architecture to decode the prompt. Additionally, we introduce a scale replay training mechanism, which effectively mitigates the scale gap from images of different resolutions. We conduct comprehensive experiments on synthetic, real-world, and fluorescence imaging datasets, showcasing the remarkable effectiveness of Prompt-SID.
\end{abstract}
\begin{links}
\link{Code}{https://github.com/huaqlili/Prompt-SID.}
\end{links}
%
\section{Introduction}
\begin{figure}[h]
    \centering
        \begin{subfigure}{0.54\linewidth}
        \centering
        \includegraphics[width=\textwidth]{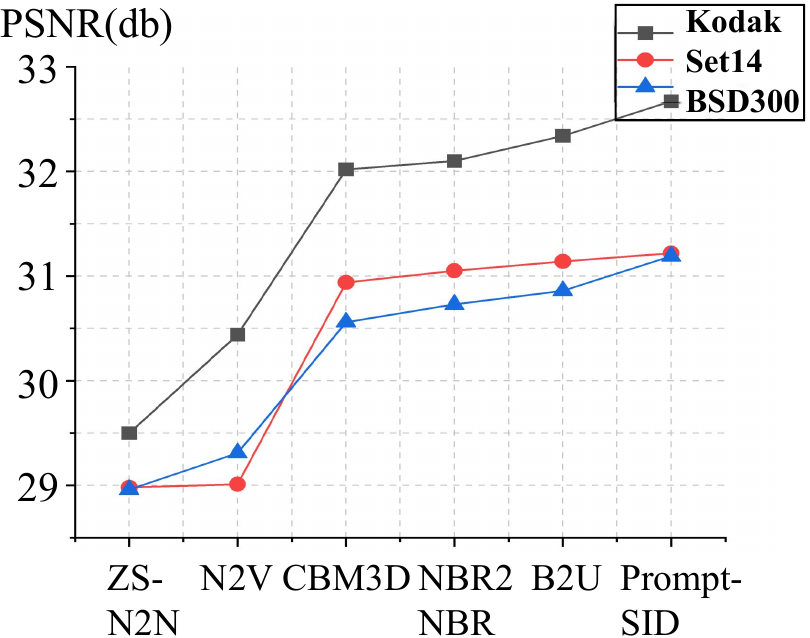}
        \caption{Performance}
    \end{subfigure}%
    \begin{subfigure}{0.46\linewidth}
        \centering
        \includegraphics[width=\textwidth]
        {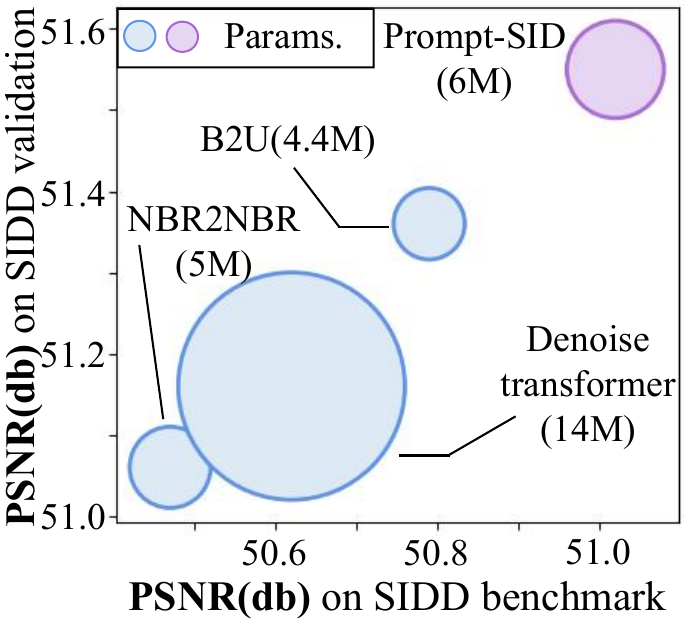}
        \caption{Parameter size}
    \end{subfigure}\\
    \caption{Comparison of Prompt-SID with other self-supervised image denoising methods in terms of model parameters and experimental results of setting $\sigma \in$ [5,50].}
    \label{fig:para}
\end{figure}
Image noise arises from diverse sources, including sensor noise and environmental factors, alongside potential introduction during quantization and image processing procedures, thereby exerting adverse impacts on downstream tasks such as classification~\cite{wang2017residual}, detection~\cite{shijila2019simultaneous}, segmentation~\cite{liu2020connecting}, and vision language model~\cite{huang2025adaptive,huang2026does}. Consequently, the quest for efficacious image denoising methodologies assumes critical significance within the domain of computer vision research.

In recent years, there has been a proliferation of learning-based supervised denoising methodologies~\cite{zhang2018ffdnet,anwar2019real,mentecs2021re,zhang2017dncnn,zamir2022learning,zamir2021multi,zhang2023kbnet}. Nonetheless, supervised denoising methods are beset by certain limitations, including their reliance on labeled data and their limited adaptability to real-world scenarios. 

Alternative paradigms such as unsupervised and self-supervised methods~\cite{laine2019high,wu2020unpaired,pang2021recorrupted,papkov2023swinia,wang2023lg,zhang2023self,lehtinen2018noise2noise} have emerged to circumvent these constraints.
Traditional self-supervised denoising methods often employ mask strategies~\cite{huang2021neighbor2neighbor,wang2022blind2unblind} to extract downsampled images or introduce blind spots by altering convolutional kernel visibility~\cite{lee2022ap,song2020denoising,krull2019noise2void}. However, these methods, although
effective in building image denoising pipelines, suffer from significant pixel information loss. During the process of sampling sub-images, some pixels are selected while others are discarded. Additionally, in the training of blind-spot networks, the central pixel of the convolutional kernel is also invisible. Furthermore, compared to blind-spot networks, downsampled images suffer from more severe structural damage and semantic degradation.

To address the aforementioned issues, we introduce Prompt-SID, a prompt-learning-based single-image denoising framework that primarily addresses the semantic degradation and structural damage caused by the sampling processes of previous self-supervised methods. We design a structural representation generation diffusion (RG-Diff) based on a latent diffusion model, using the degraded structural representations as conditional information to guide the recovery of undamaged ones. In this process, we encode information from all pixels, thereby preserving the previously invisible pixels while avoiding identity mapping. We also design a structural attention module (SAM) in the denoiser to integrate the structural representation as a prompt, further decoding them into feature images. Our approach leverages a multi-scale alternating training regime to mitigate the issues of information loss and structural disruption typically encountered during the sub-sampling process. Additionally, through the scale replay mechanism, our method effectively reduces the scale gap and achieves domain adaptation. During inference, our framework generalizes seamlessly to denoising tasks on original scale images, maintaining the integrity of structural details. Notably, just as shown in Fig. \ref{fig:para}, our approach has demonstrated impressive results on synthetic, real-world, and fluorescence imaging datasets while maintaining a relatively low parameter count.

The contributions of our work can be summarized as:
\begin{itemize}
\item Based on prompt learning, we develop a self-supervised image denoising pipeline, extracting structural representations from the original images to inform and guide the restoration process of the downsampled inputs.
\item To bridge the scale gap between the downsampled domain and the original resolution domain, we devised a branch dedicated to processing the original resolution, indirectly contributing to the optimization process to prevent pixel identity mapping.
\item Pioneering of applying diffusion models to self-supervised image denoising, we have engineered a novel structural representation generation diffusion, leveraging the powerful capability of the generative model to refine semantic representation prompts within the latent space.
\item Our method surpasses existing SOTA approaches across various datasets, including synthetic, real-world, and fluorescence imaging datasets, demonstrating its superiority in image-denoising tasks.
\end{itemize}
\begin{figure}[t]
    \centering
    \includegraphics[width=0.9\linewidth]{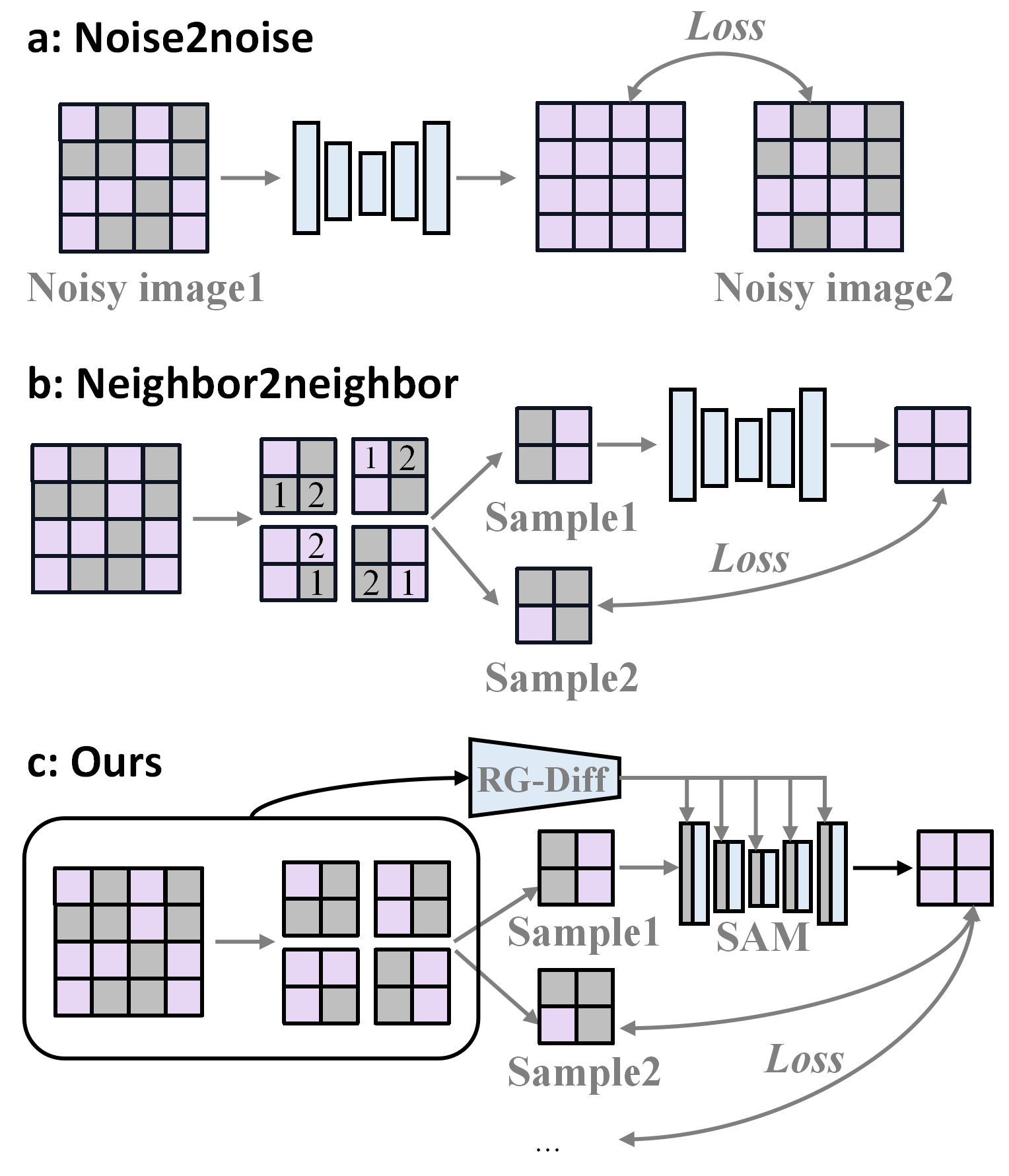}
    \caption{The distinctions between the pipelines of N2N, NBR2NBR, and Prompt-SID. }
    \label{fig:contrast}
\end{figure}

\section{Related Works}
\subsection{Self-Supervised Image Denoising}
Self-supervised image denoising methods have evolved primarily along two paths. The first path, exemplified by methods like noise2void (N2V)~\cite{krull2019noise2void}, employs blind spot to introduce invisible pixels within the central region of convolutional kernels, thereby circumventing the issue of identity mapping. Recent advancements such as AP-BSN~\cite{lee2022ap} extend blind spot networks by introducing a shuffling mechanism to disrupt the spatial continuity of noise in natural images. Additionally, some studies~\cite{wang2023lg} modify the blind spot areas within convolutional kernels. The second path, as Fig. \ref{fig:contrast} shows, 
represented by noise2noise (N2N)~\cite{lehtinen2018noise2noise}, positing that training with L2 loss tends to converge towards the mean of observed values. This suggests the feasibility of replacing desired training targets with distributions having similar means. Subsequent endeavors~\cite{huang2021neighbor2neighbor,mansour2023zero,li2023spatial} have been dedicated to self-supervised training by downsampling inputs and targets from single noisy images. 
\begin{figure*}[h]
  \centering
  \includegraphics[width=\textwidth]{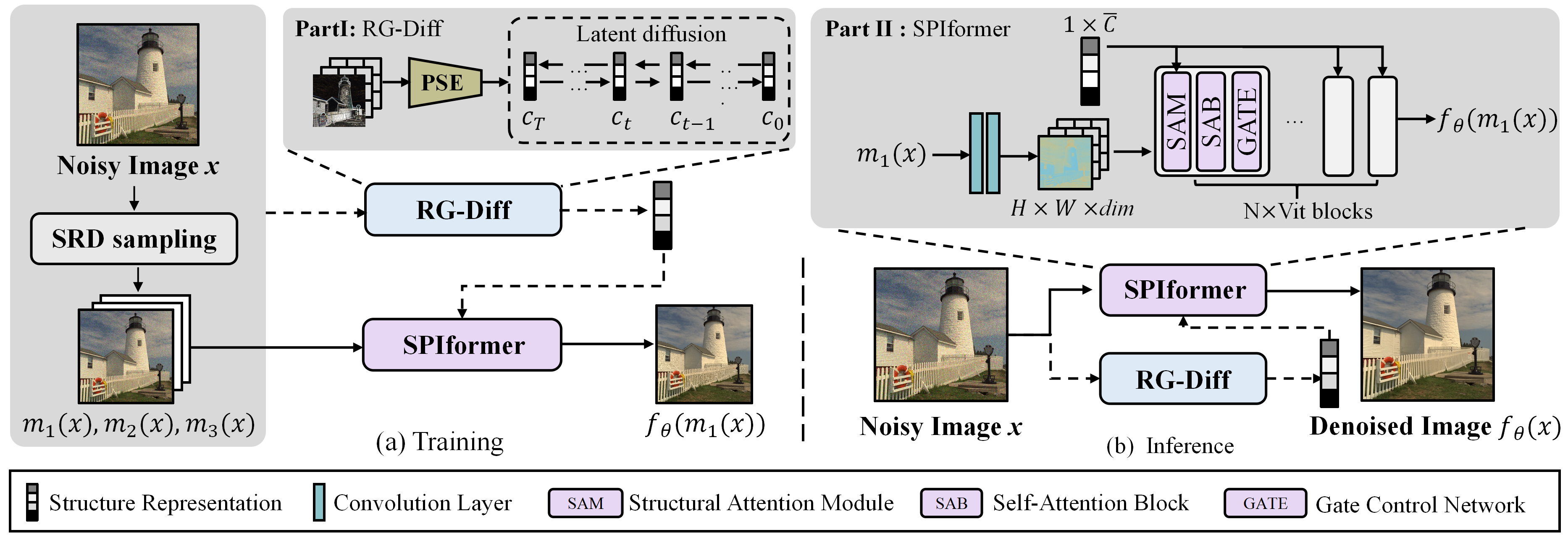}
  \caption{The primary denoising pipeline of Prompt-SID. (a) This method acquires the sub-images for network training through a spatial redundancy sampling strategy. These inputs are denoised using SPIformer, while the original image's structural representation is obtained as a prompt through RG-Diff. Each Transformer block incorporates a SAM to facilitate feature fusion. (b) During inference, Prompt-SID exclusively employs the original scale image through SPIformer and the RG-Diff branch.}
  \label{fig:pipe}
\end{figure*}
\subsection{Diffusion Model in Low-Level Tasks}

Diffusion models~\cite{ho2020denoising,song2020denoising,kawar2022denoising,huang2026real,zhang2026heterogeneous} leverage parameterized Markov chains to optimize the lower variational bound on the likelihood function, thereby enabling them to generate more precise target distributions compared to other generative models such as GANs. Over recent years, they have garnered significant attention in image restoration tasks, including super-resolution~\cite{li2022srdiff,lin2023diffbir}, enhancement~\cite{wang2024zero}, inpainting~\cite{xia2023diffir}, and so forth. 
These approaches often fine-tune a pre-trained stable diffusion model and directly decode the generated latent space representations to obtain outputs. However, these generation methods inevitably introduce a degree of randomness, resulting in subtle semantic deviations at the image level due to sampling Gaussian noise. Additionally, it fails to meet the requirements for lightweight deployment.
\section{Method}
To address the issues of low pixel utilization and structural damage, we made the following improvements in Prompt-SID. Firstly, We applied a spatial redundancy sampling strategy to minimize pixel wastage. Secondly, during the training phase at the downscaled image, we introduced RG-Diff for extracting structural representations via latent diffusion. Leveraging the generative capacity of the diffusion model, We aim for the model to utilize structural information from the downscaled images to recover corresponding representations at the original scale. The structural representations generated are then fused into the SPIformer using the SAM mechanism. Additionally, to ensure the trained model generalizes effectively on original-scale images, we incorporated a scale replay mechanism during training: following the processing of downscaled images in each iteration, gradients are frozen, and an additional inference pass is conducted on the original-scale images. The pipeline of our method is illustrated in Fig. \ref{fig:pipe}.

\subsection{Spatial Redundancy Sampling Strategy}
Following the principles of noise2noise, targets that adhere to a zero-mean noise while similar to the ground truth can serve as a supervisory signal. So we sample the input and target for network training within a single noisy image.

By employing spatial redundancy sampling strategy $m$, we can obtain sub-images $m_1(\mathbf{x})$, $m_2(\mathbf{x})$, $m_3(\mathbf{x})$ from the original-scale noisy image $x$. First, we divide the image $\mathbf{x}$ into $h$/2 $\times$ $w$/2 small blocks, with each block containing four pixels. The small block located in the i-th row and j-th column is named $b(\mathbf{i}, \mathbf{j})$. From each block, we randomly sample three adjacent pixels $p_1(b(\mathbf{i}, \mathbf{j}))$, $p_2(b(\mathbf{i}, \mathbf{j}))$, $p_3(b(\mathbf{i}, \mathbf{j}))$, where $p_1(b(\mathbf{i}, \mathbf{j}))$ is adjacent to the other two pixels. The selection of $p_2(b(\mathbf{i}, \mathbf{j}))$ and $p_3(b(\mathbf{i}, \mathbf{j}))$ is random among the remaining two pixels. Subsequently, we obtained three sub-images that are one-fourth the size of the original image. 

The process can be written as follows:
\begin{equation}
    m_n(\mathbf{x})=\sum_{\mathbf{i}=1}^{h/2} \sum_{\mathbf{j}=1}^{w/2} p_n(b(\mathbf{i},\mathbf{j})), n=1,2,3
\end{equation}
\begin{figure*}[h]
  \centering
  \includegraphics[width=\textwidth]{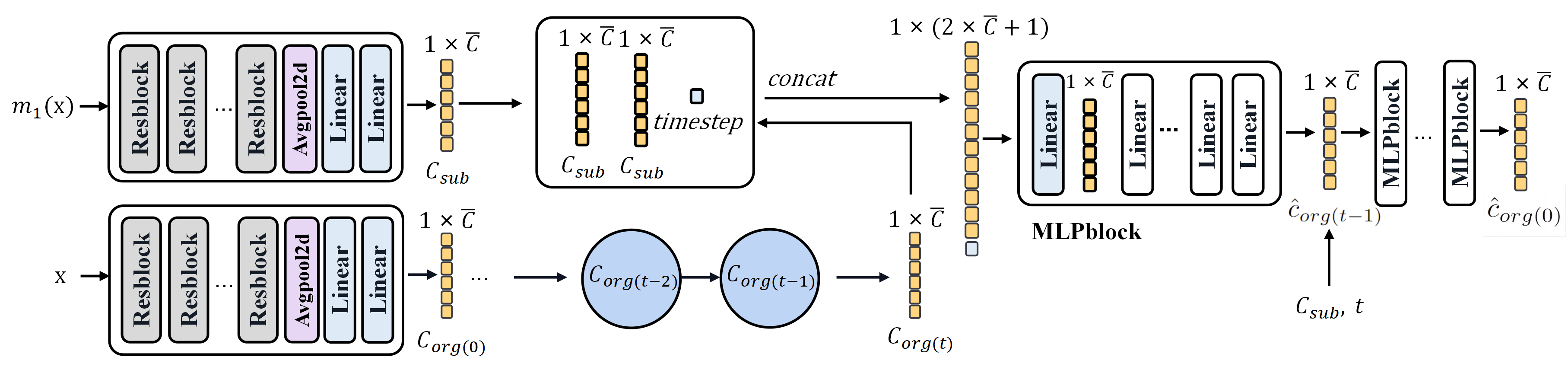}
  \caption{Diagram of the RG-Diff branch. Initially, PSE encodes the image representation into an implicit space, followed by a diffusion process within this space to obtain $\mathbf{c}_{org(t)}$. Utilizing the representation of $m_1(\mathbf{x})$ as a conditioning factor, RG-Diff guides the restoration of the representation of $\mathbf{x}$. This is achieved by merging $\mathbf{c}_{org(t)}$, $\mathbf{c}_{sub}$ and timestep $\mathbf{t}$ in the reverse diffusion stage inputting them into the denoising network.}
  \label{fig:diff}
\end{figure*}
\subsection{Structural Representation Generation Diffusion}

We propose structural representation generation diffusion (RG-Diff), performing the diffusion process within a $1\times N$ dimensional vector space. To minimize the randomness in the generation process, we design a joint training framework using the $\mathcal{L}_1$ loss in vector space, and integrating the generated representations into the feature map processing branch, rather than directly decoding them into output results. The operational principle of RG-Diff is illustrated in Fig. \ref{fig:diff}.

First, we designed a pixel structure encoder (PSE) to compress image information into the implicit space and extract structural representations. The PSE comprises several residual blocks, a global average pooling layer, and two linear layers. We encode the downscaled image \( m_1(\mathbf{x}) \) and the original scale image \( \mathbf{x} \), resulting in the structural representations of the downsampled image \( \mathbf{c}_{sub} \) and the original scale image \(mathbf{c}_{org(0)} \), respectively. The process can be represented by the following equation:
\begin{equation}
    \mathbf{c}_{sub}=PSE(m_1(\mathbf{x}))
\end{equation}
\begin{equation}
    \mathbf{c}_{org(0)}=PSE(\mathbf{x})
\end{equation}

Subsequently, we perform the forward diffusion process based on \( \mathbf{c}_{org(0)} \). At a sampled time step \( t \), the forward diffusion is carried out using the following equation, where \( \mathbf{c}_{org(0)} \) serves as the initial state. We introduce noise to this representation according to the Markov process.
\begin{equation}
    q(\mathbf{c}_{org(t)}|\mathbf{c}_{org(0)})=\mathcal{N}(\mathbf{c}_{org(t)};\sqrt{\bar{\alpha}_t}\mathbf{c}_{org(0)};(1-\bar{\alpha}_t)\mathbf{I})
\end{equation}
Here, \( \mathbf{c}_{{org(t)}} \) represents the state with noise obtained after \( \mathbf{t} \) steps of sampling on \( \mathbf{c}_{{org(0)}} \). $\bar{\alpha}_t$ is a manually designed hyperparameter. $\beta_t$ is the predefined scale factor, which increases linearly with the time steps. The relationship between $\bar{\alpha}_t$ and $\beta_t$ satisfies: $\alpha_t=1-\beta _t,
\bar{\alpha }_t= {\textstyle \prod_{i=1}^{t}}\alpha_t$.

In the reverse process, we incorporate \( \mathbf{c}_{sub} \) as a conditional control input through concatenation during the \( \mathbf{t} \)-step denoising procedure. Given that the features are one-dimensional vectors, we employ MLP for this task. Unlike the conventional reverse diffusion process, where the inputs to the denoising network consist of time step \( \mathbf{t} \) and the intermediate state \( \mathbf{c}_t \), we concatenate \( \mathbf{c}_{sub} \) with \( \mathbf{c}_{org(t)} \) at the feature level. This joint input is crucial for guiding subsequent generation steps.
The reverse diffusion process at each time step can be articulated as follows:
\begin{small}
\begin{equation}
    \hat{\mathbf{c}}_{org(t-1)}=\frac{1}{\sqrt{\alpha _t} }\{\hat{\mathbf{c}}_{org(t)}-f_\theta(\hat{\mathbf{c}}_{org(t)},\mathbf{c}_{sub},\mathbf{t})\frac{1-\alpha _t}{\sqrt{1-\bar{\alpha }_t } } \} 
\end{equation}
\end{small}
In this context, \( f_\theta \) denotes the parameters of the denoising network. Due to the computational efficiency of this branch, we perform reverse diffusion across all time steps to obtain the final representation \( \hat{\mathbf{c}}_{org(0)} \). To control the generation direction, we impose the following constraint using $L_1$ loss.
\begin{equation}
    \mathcal{L}_{diff}=\left \| \hat{\mathbf{c}}_{org(0)}-\mathbf{c}_{org(0)} \right \|_1 
\end{equation}
The structural representation \( \hat{\mathbf{c}}_{org(0)} \) is utilized in the image restoration branch for decoding, thereby directing the generation at the feature map level.
\begin{figure}[t]
  \centering
  \includegraphics[width=\linewidth]{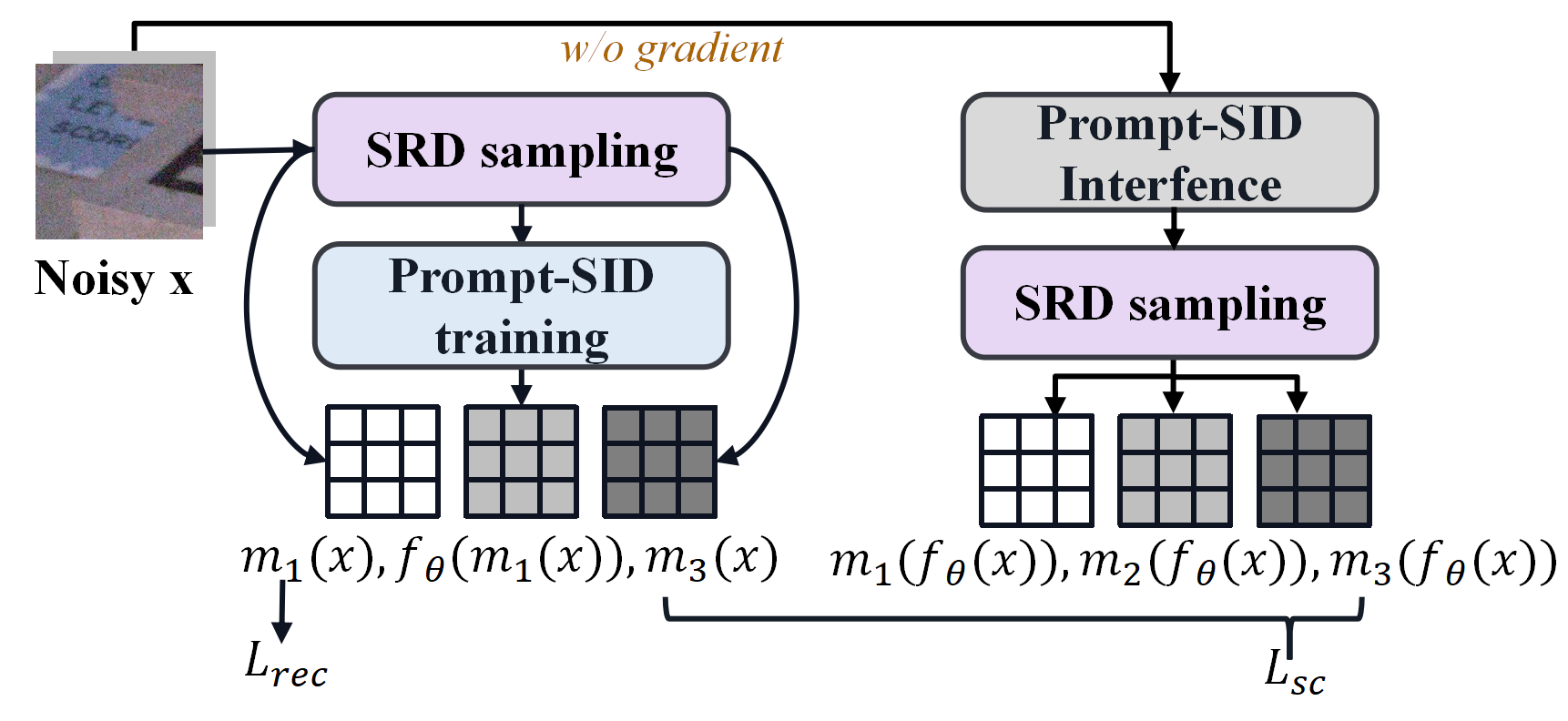}
  \caption{Introducing a scale-replay training branch without gradient backpropagation. We pass the original-scale noisy image $\mathbf{x}$ through Prompt-SID and downsample the denoised result to obtain $m_1(f_\theta(\mathbf{x}))$, $m_2(f_\theta(\mathbf{x}))$, $m_3(f_\theta(\mathbf{x}))$. These downscaled outputs are utilized to enforce regularization constraints on the image restoration loss.}
  \label{fig:loss}
\end{figure}
\subsection{Structural Prompt Integrative Transformer}
We employ the vision transformer (ViT)~\cite{dosovitskiy2020image} module as the branch of image reconstruction. Similar to prior research~\cite{zamir2022restormer,zhang2023kbnet,chen2022simple}, our transformer module comprises two components: the multi-head self-attention block and the gate control network. Additionally, we introduce a structural attention module (SAM) to incorporate the previously generated structural representation \( \hat{\mathbf{c}}_{org(0)} \) into the feature map. 

The primary operation principle of SAM can be delineated into two phases: channel attention extraction and computation, and the integration of structural embedding information. We acquire channel attention weights through global average pooling and 1x1 convolution applied to the feature maps, as illustrated by the following equation:
\begin{equation}
    \mathbf{c}_{sca}=AvgPool(\hat{\mathbf{F}}) * \mathbf{W}_{l1} + \mathbf{b}_{l1}
\end{equation}
In this equation, \( AvgPool(\hat{\mathbf{F}}) \) denotes the global average pooling operation applied to the feature map \( \hat{\mathbf{F}} \). The resulting output is subsequently multiplied by the weight \( \mathbf{W}_{l1} \) and added to the bias \( \mathbf{b}_{l1} \). Subsequently, we merge \( \mathbf{c}_{sca} \) and \( \hat{\mathbf{c}}_{org(0)} \) to jointly derive channel attention weights that direct the processing of the feature maps. The precise procedure is outlined as follows:
\begin{equation}
\begin{split}
\mathcal{F}&=\mathbf{W}_{s1}\mathbf{c}_{sca}\odot \mathbf{W}_{c1}\hat{\mathbf{c}}_{org(0)}\odot Norm(\hat{\mathbf{F}})\\&+\mathbf{W}_{s2}\mathbf{c}_{sca}\odot \mathbf{W}_{c2}\hat{\mathbf{c}}_{org(0)}
\end{split}
\end{equation}
In the equation mentioned above, $\mathbf{W}_{s1},\mathbf{W}_{s2},\mathbf{W}_{c1},\mathbf{W}_{c2}$ represents the weight matrix of the linear layer. $\mathcal{F}$ represents the feature map processed by the SAM.

\begin{figure*}[t]
    \centering
    \begin{subfigure}{0.5\textwidth}
        \centering
        \includegraphics[width=\textwidth]{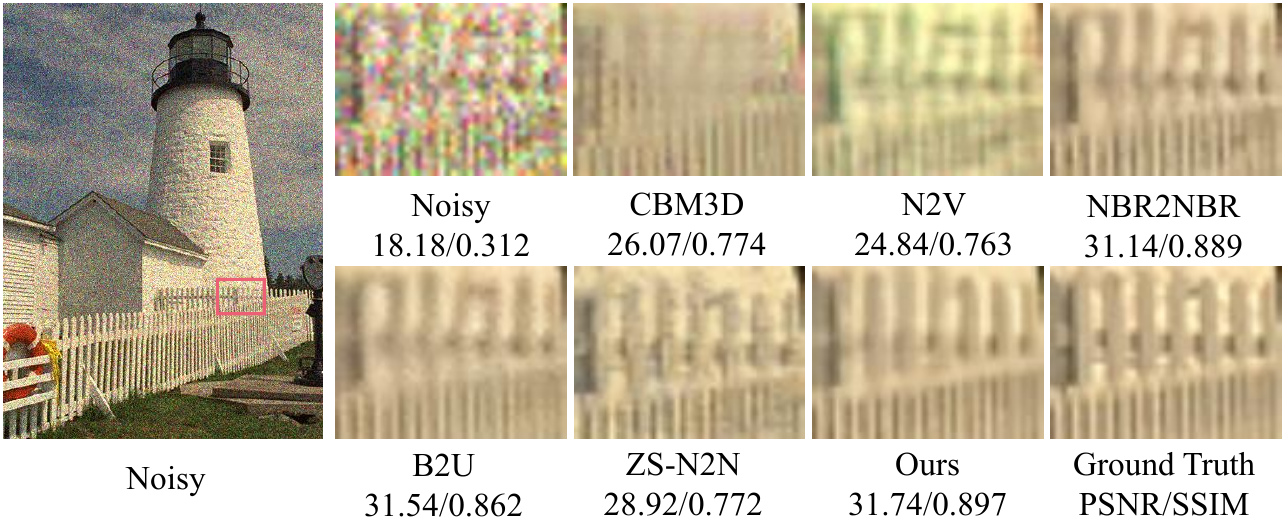}
    \end{subfigure}%
    \begin{subfigure}{0.5\textwidth}
        \centering
        \includegraphics[width=\textwidth]{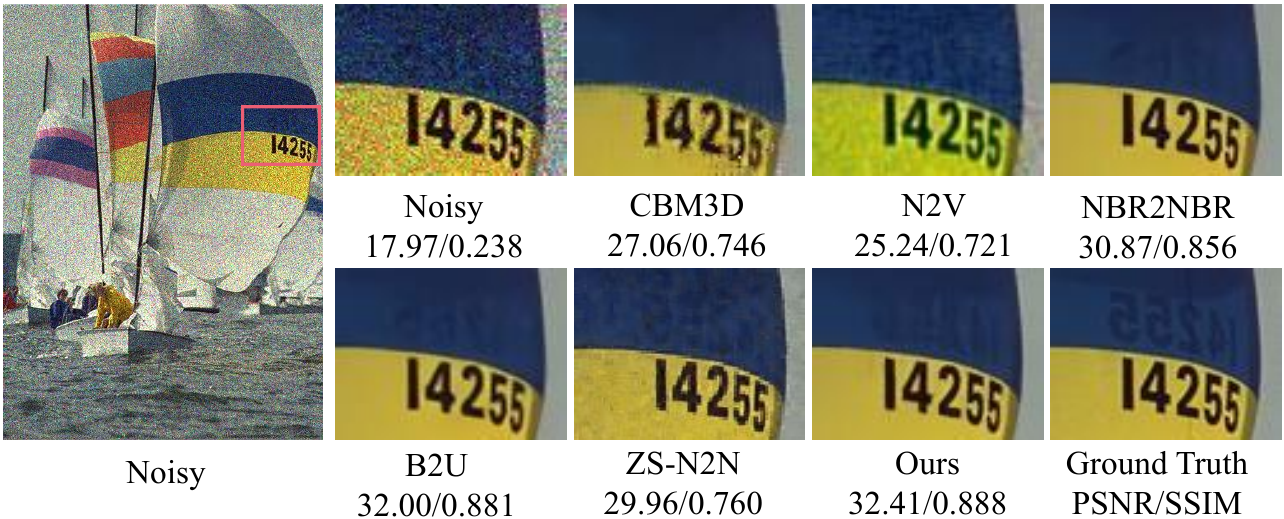}
    \end{subfigure}
    \caption{The visual comparison of Prompt-SID
     with state-of-the-art self-supervised image denoising methods in synthetic noise experiments, demonstrating results for Poisson noise set at level 30 tested on the Kodak dataset.}
    \label{fig:srgb}
\end{figure*}
\begin{table*}[h]\small
\renewcommand\arraystretch{1.1}
\centering
\begin{tabular}{ccccccccc}
\specialrule{1pt}{0pt}{0pt}
   &Dataset& \multicolumn{1}{c|}{Baseline,N2C}&CBM3D&N2V&NBR2NBR&B2U&ZS-N2N&Ours                \\ \hline
\multirow{3}{*}{$\sigma$ = 25}   & Kodak                                        & \multicolumn{1}{c|}{32.43/0.884}  & 31.87/0.868 & 30.32/0.821 & 32.08/0.879 & 32.27/0.880 & 29.25/0.779 & \textbf{32.41/0.883} \\
                           & BSD300                                       & \multicolumn{1}{c|}{31.05/0.879}  & 30.48/0.861 & 29.34/0.824 & 30.79/0.873 & 30.87/0.872 & 28.56/0.801 & \textbf{31.16/0.880} \\
                           & Set14                                         & \multicolumn{1}{c|}{31.40/0.869}  & 30.88/0.854 & 28.84/0.802 & 31.09/0.864 & 31.27/0.864 & 28.29/0.779 & \textbf{31.45/0.868} \\ \hline
\multirow{3}{*}{$\sigma \in$ [5,50]} & Kodak                                        & \multicolumn{1}{c|}{32.51/0.875}  & 32.02/0.860 & 30.44/0.806 & 32.10/0.870 & 32.34/0.872 & 29.50/0.767 & \textbf{32.67/0.876} \\
                           & BSD300                                       & \multicolumn{1}{c|}{31.07/0.866}  & 30.56/0.847 & 29.31/0.801 & 30.73/0.861 & 30.86/0.861 & 28.96/0.797 & \textbf{31.19/0.866} \\
                           & Set14                                         & \multicolumn{1}{c|}{31.41/0.863}  & 30.94/0.849 & 29.01/0.792 & 31.05/0.858 & 31.14/0.857 & 28.98/0.783 & \textbf{31.22/0.860} \\ \specialrule{1pt}{0pt}{0pt}
\end{tabular}
\caption{Quantitative results on synthetic datasets in sRGB space for Gaussian noise set. The highest PSNR(dB)/SSIM among unsupervised denoising methods are highlighted in \textbf{bold}.}
\label{tab:syn1}
\bigskip
\centering
\begin{tabular}{ccccccccc}
\specialrule{1pt}{0pt}{0pt}
   &Dataset& \multicolumn{1}{c|}{Baseline,N2C}&CBM3D&N2V&NBR2NBR&B2U&ZS-N2N&Ours                \\ \hline
\multirow{3}{*}{$\lambda$ = 30}   & Kodak                                        & \multicolumn{1}{l|}{31.78/0.876}  & 30.53/0.856                  & 28.90/0.788             & 31.44/0.870                 & 31.64/0.871             & 28.70/0.756                & \textbf{31.65/0.874}     \\
                             & BSD300                                       & \multicolumn{1}{l|}{30.36/0.868}  & 29.18/0.842                  & 28.46/0.798             & 30.10/0.863                 & 30.25/0.862             & 28.07/0.787                & \textbf{30.43/0.869}     \\
                             & Set14                                        & \multicolumn{1}{l|}{30.57/0.858}  & 29.44/0.837                  & 27.73/0.774             & 30.29/0.853                 & 30.46/0.852             & 27.72/0.758                & \textbf{30.56/0.858}     \\ \hline
\multirow{3}{*}{$\lambda \in$ [5,50]} & Kodak                                        & \multicolumn{1}{l|}{31.19/0.861}  & 29.40/0.836                  & 28.78/0.758             & 30.86/0.855                 & 31.07/0.857             & 28.10/0.725                & \textbf{31.49/0.864}     \\
                             & BSD300                                       & \multicolumn{1}{l|}{29.79/0.848}  & 28.22/0.815                  & 27.92/0.766             & 29.54/0.843                 & 29.92/0.852             & 27.68/0.765                & \textbf{30.01/0.855}     \\
                             & Set14                                        & \multicolumn{1}{l|}{30.02/0.842}  & 28.51/0.817                  & 27.43/0.745             & 29.79/0.838                 & 30.10/0.844             & 27.51/0.748                & \textbf{30.34/0.852}     \\ \specialrule{1pt}{0pt}{0pt}
\end{tabular}
\caption{Quantitative results on synthetic datasets in sRGB space for Poisson noise set. The highest PSNR(dB)/SSIM among unsupervised denoising methods are highlighted in \textbf{bold}.}
\label{tab:syn2}
\end{table*}
\subsection{Scale Replay Mechanism and Loss}
After passing through SPIformer, we derive \( f_\theta(m_1(\mathbf{x})) \), where \( f_\theta \) denotes the network parameters requiring optimization in RG-Diff and SPIformer. The reconstruction loss is computed by evaluating the $L_2$ loss between \( f_\theta(m_1(\mathbf{x})) \) and \( m_2(\mathbf{x}) \), as well as between \( f_\theta(m_1(\mathbf{x})) \) and \( m_3(\mathbf{x}) \). The specific formula is outlined as follows:
\begin{small}
\begin{equation}
    \mathcal{L}_{rec}\!=\!\left \| f_\theta(m_1(\mathbf{x}))\!-\!m_2(\mathbf{x}) \right \|_2 \!+\!\left \| f_\theta(m_1(\mathbf{x}))\!-\!m_3(\mathbf{x}) \right \|_2
\end{equation}
\end{small}

In the preceding discussion, we emphasized the necessity of addressing the generalization between downscaled and original-scale images. Our objective is to train a model capable of alleviating the domain gap between them. Therefore, in each iteration, we conduct an additional inference process on the original-scale images. The steps of the model inference process are illustrated in Fig. \ref{fig:pipe}. We encode \( \mathbf{x} \) using the PSE and feed the structural representation $\mathbf{c}_x$ into RG-Diff, which introduces random Gaussian noise during inference and performs reverse diffusion guided by \( \mathbf{c}_x \). Concurrently, \( \mathbf{x} \) undergoes processing through a feature manipulation branch where structural representations are fused.

The overall training process with the scale replay mechanism is illustrated in Fig. \ref{fig:loss}. To prevent identity mapping, we compute losses using the downsampled version of the denoised original-scale image, rather than directly supervised by the noisy original image.
\begin{small}
\begin{equation}
\begin{split}
    \mathcal{L}_{sc} &\!=\!\left \| f_\theta(m_1(\mathbf{x}))\!-m_1(f_\theta(\mathbf{x}))\!-m_2(\mathbf{x})\!+\!m_2(f_\theta(\mathbf{x})) \right \|_2 \\&\! +\!\left \| f_\theta(m_1(\mathbf{x}))\!-\!m_1(f_\theta(\mathbf{x}))-m_3(\mathbf{x})\!+\!m_3(f_\theta(\mathbf{x}))\right \|_2
\end{split}
\end{equation}
\end{small}
The expression for the final loss is as follows:
\begin{equation}
\mathcal{L}=\alpha_{rec}\mathcal{L}_{rec}+\alpha_{sc}\mathcal{L}_{sc} +\alpha_{diff}\mathcal{L}_{diff}
\end{equation}
\( \alpha_{rec} \), \( \alpha_{sc} \) and \( \alpha_{diff} \) are adjustable hyperparameters. In our experiments, we set them to 1, 1.5, and 1, respectively.

\begin{figure*}[h]
  \centering
  \includegraphics[width=\textwidth]{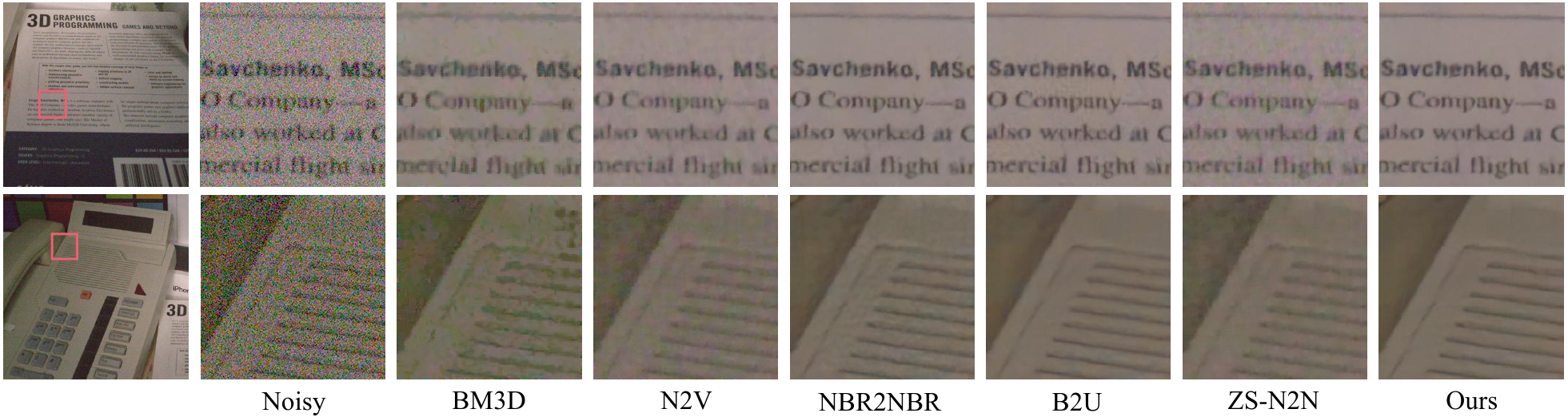}
  \caption{Visual comparison of our method against other methods on SIDD Benchmark. All images are converted from raw RGB space to sRGB space by the ISP provided by SIDD for visualization. Best viewed in color.}
  \label{fig:sidd}
\end{figure*}

\begin{table*}[]\small
\renewcommand\arraystretch{1.1}
\centering
\resizebox{\textwidth}{!}{
\begin{tabular}{ccccccccc}
\specialrule{1pt}{0pt}{0pt}
SIDD dataset& \multicolumn{1}{c|}{Baseline,N2C}&CBM3D&N2V&NBR2NBR&B2U&ZS-N2N&DT&Ours                \\ \hline
Benchmark           & \multicolumn{1}{c|}{50.60/0.991}  & 48.60/0.986 & 48.01/0.983 & 50.47/0.990 & 50.79/0.991 & 47.68/0.981 &50.62/0.990 & \textbf{51.02/0.991} \\
Validation          & \multicolumn{1}{c|}{51.19/0.991}  & 48.92/0.986 & 48.55/0.984 & 51.06/0.991 & 51.36/0.992 & 48.75/0.985 &51.16/0.991 & \textbf{51.55/0.992}\\ \specialrule{1pt}{0pt}{0pt}
\end{tabular}
}
\caption{Quantitative comparisons (PSNR(dB)/SSIM) on
SIDD benchmark and validation datasets in raw-RGB
space. The best PSNR(dB)/SSIM results among denoising
methods are marked in \textbf{bold}.}
\label{tab:sidd}
\end{table*}
\section{Experiment}
\subsection{Implementation Details}
\textbf{Training Details.} We select supervised method~\cite{ronneberger2015u}, CBM3D~\cite{dabov2007color}, BM3D~\cite{dabov2007image}, anscombe~\cite{makitalo2010optimal},  noise2void(N2V)~\cite{krull2019noise2void}, NBR2NBR~\cite{huang2021neighbor2neighbor}, blind2unblind(B2U)~\cite{wang2022blind2unblind}, zero shot noise2noise(ZS-N2N)~\cite{mansour2023zero} for writing. More comparative experimental results can be found in the supplementary material. We obtain quantitative and qualitative results from other methods by adopting official pre-trained models and running their public codes. For training, we fixed the decay rate for the exponential moving average at 0.999 and initialized the learning rate to 0.0002. Parameter optimization and computation were performed with Adam optimizer, setting \( \beta_1 \) to 0.9 and \( \beta_2 \) to 0.99. All training was executed on one Nvidia RTX3090.

\noindent\textbf{Datasets.} In synthetic denoising, we curated a training set comprising 44,328 images from the ILSVRC2012 dataset~\cite{deng2009imagenet}, with testsets named kodak, BSD300~\cite{martin2001database}, and set14~\cite{zeyde2012single}. In real-world denoising, We utilized SIDD-Medium dataset~\cite{abdelhamed2018high} in the raw-RGB domain as the training set. For testing, we employed two datasets: SIDD validation and SIDD benchmark. Moreover, We employed the two-photon calcium imaging of a large 3D neuronal populations dataset, as proposed by SRDtrans~\cite{li2023spatial}, for training and testing on the fluorescence imaging dataset. More implementation details can be found in the supplementary material.
\subsection{Benchmarking Results}
\noindent\textbf{Synthetic Denoising.}
For sRGB images, we conducted four sets of experiments under Gaussian and Poisson noise settings. Our results are displayed in Fig. \ref{fig:srgb}, Tab. \ref{tab:syn1} and \ref{tab:syn2}. Overall, our method attained outstanding results, exhibiting superior performance across the majority of experimental metrics. Specifically, in all experimental trials, our approach surpassed SOTA method B2U~\cite{wang2022blind2unblind} on all test datasets. Furthermore, our method exhibited a consistent 0.21-0.34dB improvement over another sampling method NBR2NBR~\cite{huang2021neighbor2neighbor} across various metrics, owing to our structural representation prompt strategy. Notably, Our approach demonstrates measurable enhancements over traditional supervised methods~\cite{ronneberger2015u}. Specifically, in the $\lambda \in [5,50]$ experiment, we outperformed the baseline by 0.32dB on the Set14 dataset. Moreover, Across twelve experimental settings spanning three datasets, our method outperformed supervised approaches in eight instances.

\begin{table}[t]\small
\centering
\resizebox{\linewidth}{!}{
\begin{tabular}{ccccc}
\specialrule{1pt}{0pt}{0pt}
Sampling rate &\multicolumn{1}{c}{Baseline,N2C}&NBR2NBR&B2U&Ours           \\ \hline
1 Hz                            & \multicolumn{1}{c}{15.65}        & 15.18   & 15.46 & \textbf{15.89} \\
3 Hz                            & \multicolumn{1}{c}{16.28}        & 15.58   & 15.65 & \textbf{15.98} \\10 Hz                           & \multicolumn{1}{c}{16.14}        & 15.79   & 15.98          & \textbf{16.06} \\
 30 Hz                           & \multicolumn{1}{c}{20.89}        & 20.21   & \textbf{21.12} & 21.10          \\ \specialrule{1pt}{0pt}{0pt}
\end{tabular}
}
\caption{The fluorescence imaging denoising experiment's quantitative results were assessed using SNR(db). The best results among denoising
methods are marked in \textbf{bold}.}
\label{tab:cad}
\end{table}
\begin{figure}[t]
  \centering
  \includegraphics[width=\linewidth]{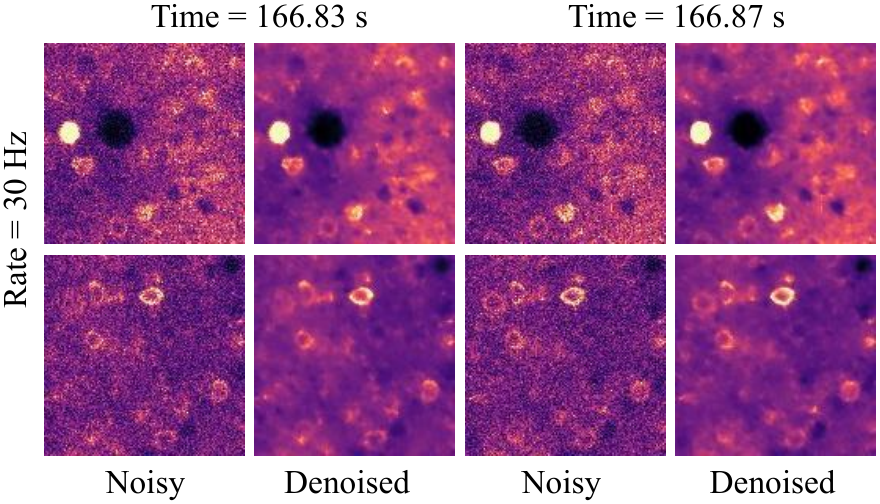}
  \caption{Visual results of fluorescence imaging datasets.}
  \label{fig:cad}
\end{figure}

\noindent\textbf{Real-world Denoising.}
The quantitative results on real-world datasets are presented in Tab. \ref{tab:sidd}. On the SIDD dataset in the raw-RGB domain, we achieved a 0.55 dB and 0.49 dB advantage on the SIDD validation and SIDD benchmark datasets respectively, relative to the original architecture of our method NBR2NBR~\cite{huang2021neighbor2neighbor}. In comparison to the previous state-of-the-art method, B2U~\cite{wang2022blind2unblind}, we demonstrated improvements of 0.23 dB and 0.19 dB. This can be attributed to the more efficient attention mechanism of the transformer compared to traditional convolutions, and it also underscores the effectiveness of the integrated re-visualization pixel strategy from the network structure. Furthermore, we surpassed the Denoise Transformer(DT)~\cite{zhang2023self} method that utilizes a transformer as the backbone. This further validates the effectiveness of diffusion in generating prompts that fuse multi-scale information. By visualizing the results in Fig. \ref{fig:sidd}, we observe that Prompt-SID outperforms in preserving details, minimizing edge blurring and color imbalance.

\noindent\textbf{Fluorescence Imaging Denoising.}
Our results on fluorescence imaging denoising are outlined in Tab. \ref{tab:cad}. For comparative analysis, we selected baseline methods N2C~\cite{ronneberger2015u}, NBR2NBR~\cite{huang2021neighbor2neighbor}, and B2U~\cite{wang2022blind2unblind}. Prompt-SID outperforms other self-supervised methods and achieves results comparable to supervised approaches. Notably, our results surpass the supervised baseline performance at both 1Hz and 30Hz scanning speeds. Upon visualizing the results in Fig. \ref{fig:cad}, we observed that our method exhibits strong generalization to fluorescence imaging data distribution and achieves remarkable image restoration even with significant noise.  
\begin{table}[]\small
\begin{tabular}{cccc}
\specialrule{1pt}{0pt}{0pt}
                          Dataset & SIDD & Kodak                & Set14                \\ \hline 
w/o RG-Diff               & 51.32/0.991            & 31.97/0.874          & 31.06/0.862          \\ 
w/o RG condition     & 51.40/0.991            & 32.25/0.881          & 31.32/0.867          \\ 
w/o $\mathcal{L}_{sc}$                & 50.97/0.990            & 32.01/0.879          & 30.89/0.861          \\ 
w/o $\mathcal{L}_{diff}$                 & 51.03/0.990            & 32.32/0.883          & 31.38/0.868          \\ 
ours                      & \textbf{51.55/0.992}   & \textbf{32.41/0.883} & \textbf{31.45/0.868} \\ \specialrule{1pt}{0pt}{0pt}
\end{tabular}
\caption{Ablation studies on the effect of different modules in real-world and Gaussian $\sigma$ = 25 datasets.}
\label{tab:ab}
\end{table}
\begin{figure}[t]
  \centering
  \includegraphics[width=\linewidth]{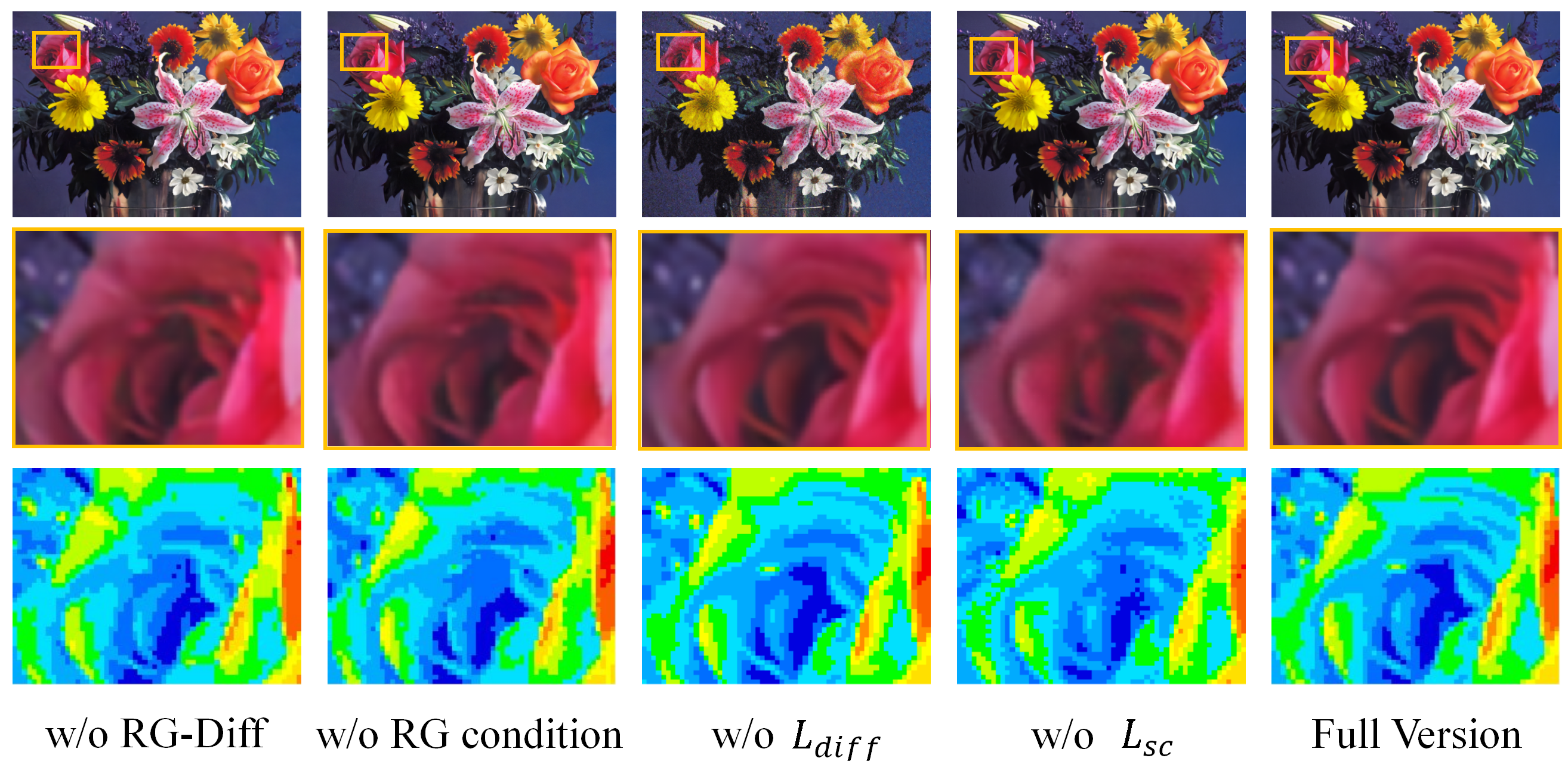}
  \caption{Visual results of ablation experiments.}
  \label{fig:ab}
\end{figure}
\begin{figure}[t]
  \centering
  \includegraphics[width=\linewidth]{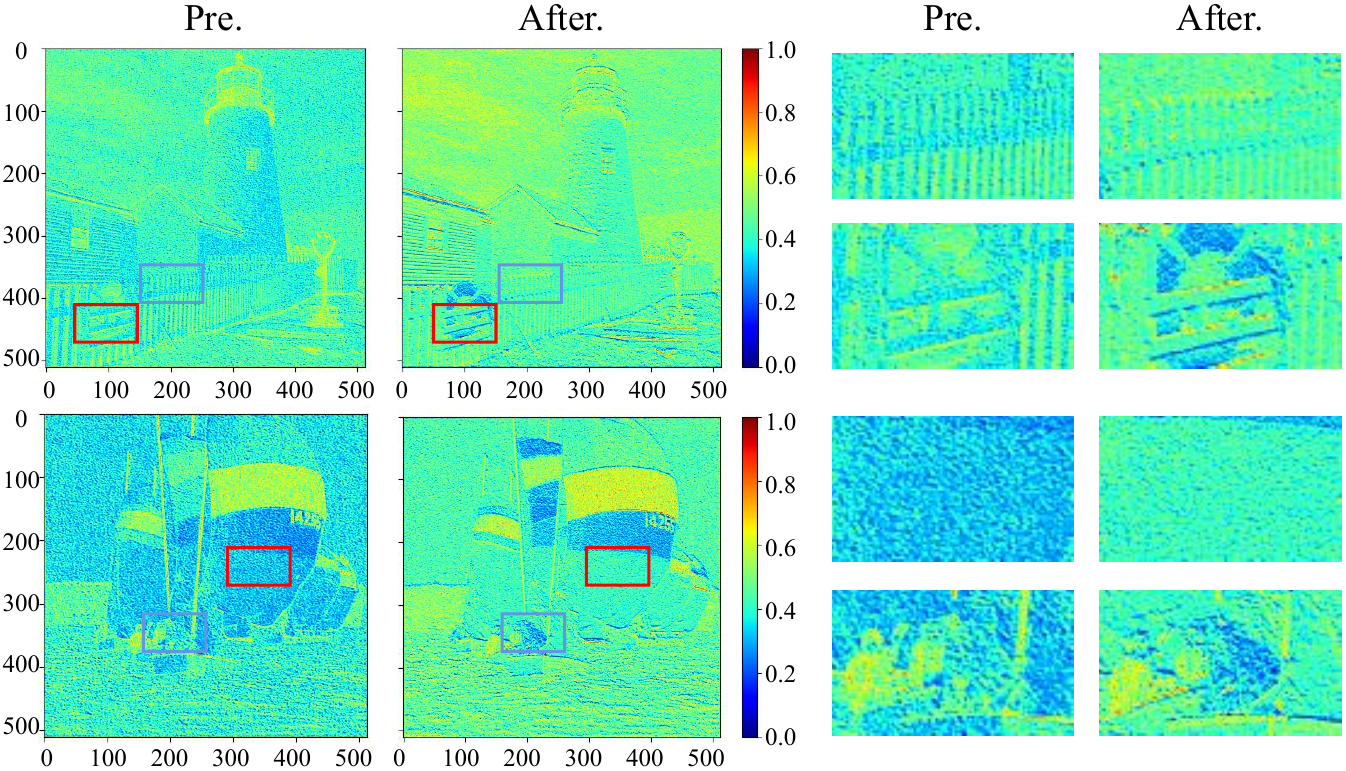}
  \caption{The visualization of the feature map, where pre. and after. represent before and after the prompt fusion.}
  \label{fig:fea}
\end{figure}
\subsection{Ablation Study}
\textbf{The Ablation of Several Modules.} We conducted the following module ablation experiments and tested them on the SIDD benchmark, as well as Kodak and Set14.

The settings for the four sets of experiments are as follows: 1) Ablation experiment on the Structural representation Prompt. We conducted an ablation on RD-Diff within Prompt-SID, removing the Structural representation Prompt, and simultaneously eliminating the fusion mechanism of SAM in the denoiser. 2)Within the RD-Diff branch, we omitted the mechanism that uses the structural representation of downsampled images as a conditioning factor for generation, substituting it with an equally shaped Gaussian noise. Consequently, the diffusion model branch transformed into a traditional unconstrained generative branch. 3)We removed the scale replay mechanism, thereby excluding its influence on model training in the loss function $\mathcal{L}_{sc}$. During training, the denoiser solely processes downsampled images and structural representation prompts. 4)We removed $\mathcal{L}_{diff}$, imposing no loss constraints on the generation of structural representations.

The experimental results are presented in Tab. \ref{tab:ab} and Fig. \ref{fig:ab}. The full version of Prompt-SID exhibits performance improvements compared to the other ablation experiments. In sets 1 and 2, there is a degradation in image semantic details (such as the stacking of petals at the top of the rose). After removing the scale replay mechanism, the denoised images are blurrier than Prompt-SID, as the model did not encounter higher resolution information during training.

\noindent\textbf{How Does the Prompt Work?} To validate the effectiveness of the structural representation mechanism, we designed an ablation experiment to visualize the feature maps before and after prompt integration. We averaged the images along the channel dimension during the first SAM operation on the feature maps. The experimental results are shown in Fig. \ref{fig:fea}. The results demonstrate that the structural representation possesses the implicit ability to restore semantic structures and high-frequency edges. In feature map computation, prompt fusion emphasizes channels with richer structural details and semantic representations while attenuating the influence of noisy channels for high-frequency filtering.
\section{Conclusion}
We present Prompt-SID, a prompt-learning-based self-supervised image denoising framework that primarily addresses the semantic degradation and structural damage caused by the sampling processes of previous self-supervised methods. Our approach demonstrates the immense potential of the diffusion model and prompt-learning in image denoising tasks. We design a structural representation generation diffusion(RG-Diff) based on a latent diffusion model, using the degraded structural representations as conditional information to guide the recovery of undamaged ones. Additionally, through the scale replay mechanism, our method effectively reduces the scale gap between sub-sampled and original scale images. Extensive experiments demonstrate that our method consistently achieves state-of-the-art performance across synthetic, real-world, and fluorescence imaging datasets.
\section{Acknowledgements}
This work is supported by the Shenzhen Science and Technology Project under Grant (JCYJ20220818101001004).

\bibliography{aaai25}

\end{document}